%% file: main.tex
\documentclass[sigconf, authorversion]{acmart}

\AtBeginDocument{%
  }

\copyrightyear{2025}
\acmYear{2025}
\setcopyright{acmlicensed}
\acmConference[MM '25] {Proceedings of the 33rd ACM International Conference on Multimedia}{October 27--31, 2025}{Dublin, Ireland.}
\acmBooktitle{Proceedings of the 33rd ACM International Conference on Multimedia (MM '25), October 27--31, 2025, Dublin, Ireland}
\acmISBN{979-8-4007-2035-2/2025/10}
\acmDOI{10.1145/3746027.3755565}

\usepackage{booktabs}
\usepackage{color}
\usepackage{multirow}
\usepackage[table]{xcolor}
\input{packages}

\begin{document}

\title[STATUS Bench: A Rigorous Benchmark for Evaluating Object State Understanding in Vision-Language Models]{STATUS Bench: A Rigorous Benchmark for Evaluating\\Object State Understanding in Vision-Language Models}

\author{Mahiro Ukai}
\affiliation{%
  \institution{Institute of Science Tokyo,}
  \city{Tokyo}
  \country{Japan}
}

\author{Shuhei Kurita}
\affiliation{%
  \institution{National Institute of Informatics,}
  \city{Tokyo}
  \country{Japan}
}
\affiliation{%
  \institution{Institute of Science Tokyo,}
  \city{Tokyo}
  \country{Japan}
}

\author{Nakamasa Inoue}
\affiliation{%
  \institution{Institute of Science Tokyo,}
  \city{Tokyo}
  \country{Japan}
}

\input{sections/00_abstract}


\maketitle

\input{sections/01_introduction}
\input{sections/02_related_work}

\input{sections/03_status}
\input{sections/04_experiments}

\input{sections/05_conclusion}

\begin{acks}
This work was supported by JST CRONOS Japan Grant Number JPMJCS24K6 and JST BOOST Program Japan Grant Number JPMJBY24C5 and was carried out using the TSUBAME4.0 supercomputer at Institute of Science Tokyo.
\end{acks}

\bibliographystyle{ACM-Reference-Format}
\bibliography{main}

\input{sections/99_supplemental}

\end{document}

%% file: packages.tex
\usepackage{tcolorbox}
\usepackage{colortbl}
\usepackage{xcolor}
\usepackage{amsmath}
\usepackage{algorithmic}
\usepackage{algorithm}
\usepackage{bm}
\usepackage{booktabs}
\usepackage{comment}
\usepackage{multirow}
\usepackage{framed}

\usepackage{url}
\makeatletter
\newcommand{\figcaption}[1]{\def\@captype{figure}\caption{#1}}
\newcommand{\tblcaption}[1]{\def\@captype{table}\caption{#1}}
\makeatother

\newcount\K
\def\bla#1{
\K=0 \loop\ifnum\K<#1
{\textcolor[gray]{0.9}{{\it bla bla bla bla bla bla bla bla bla bla bla bla bla bla bla}}}
\advance\K by1\repeat
}

\def\myparagraph#1{\noindent \textbf{#1.}}
\usepackage{lineno}
\usepackage{makecell}
\usepackage{stmaryrd}
\usepackage{pifont}

%% file: sections/00_abstract.tex
\begin{abstract}
Object state recognition aims to identify the specific condition of objects, such as their positional states (\textit{e.g.,} open or closed) and functional states (\textit{e.g.,} on or off).
While recent Vision-Language Models (VLMs) are capable of performing a variety of multimodal tasks, it remains unclear how precisely they can identify object states.
To alleviate this issue, we introduce the \textbf{STAte and Transition UnderStanding Benchmark (STATUS Bench)},
the first benchmark for rigorously evaluating the ability of VLMs to understand subtle variations in object states in diverse situations.
Specifically, STATUS Bench introduces a novel evaluation scheme that requires VLMs to perform three tasks simultaneously: object state identification (OSI), image retrieval (IR), and state change identification (SCI).
These tasks are defined over our fully hand-crafted dataset involving image pairs, their corresponding object state descriptions and state change descriptions.
Furthermore, we introduce a large-scale training dataset, namely STATUS Train, which consists of 13 million semi-automatically created descriptions.
This dataset serves as the largest resource to facilitate further research in this area.
In our experiments, we demonstrate that STATUS Bench enables rigorous consistency evaluation and reveal that current state-of-the-art VLMs still significantly struggle to capture subtle object state distinctions.
Surprisingly, under the proposed rigorous evaluation scheme, most open-weight VLMs exhibited chance-level zero-shot performance.
After fine-tuning on STATUS Train, Qwen2.5-VL achieved performance comparable 
to Gemini 2.0 Flash.
These findings underscore the necessity of STATUS Bench and Train for advancing object state recognition in VLM research.
\end{abstract}

%% file: sections/01_introduction.tex
\begin{figure}[t]
\centering
\includegraphics[width=\linewidth]{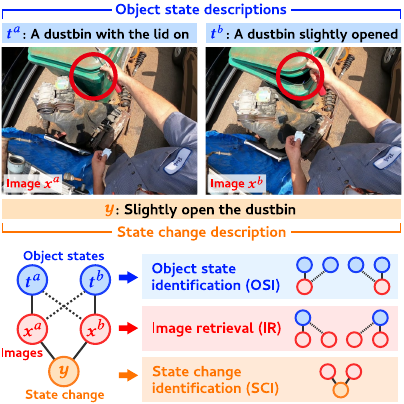}
\vspace{-14pt}
\caption{
STATUS Bench for rigorously evaluating the ability of VLMs to understand object states.
We provide quintuplets, each consisting of a image pair $\bm{(x^{a}, x^{b})}$, corresponding object state descriptions $\bm{(t^{a}, t^{b})}$ and a state change description $\bm{y}$ (top).
On the graph structure derived from these quintuplets, we define three evaluation tasks (OSI, IR, and SCI) using five sub-graphs to rigorously evaluate consistency (bottom).
}
\label{fig:top}
\end{figure}

\section{Introduction}

In recent years, vision-language models (VLMs) have made remarkable progress.
Following the success of CLIP~\cite{Radford2021CLIP}, which produces image embeddings aligned with text embeddings, numerous VLMs, such as LLaVA~\cite{liu2023LLaVA}, VILA~\cite{Lin2024VILA} and InternVL~\cite{Chen2024InternVL} have been proposed.
These VLMs are benchmarked from various perspectives, covering tasks ranging from real-world visual question answering to abstract mathematical reasoning~\cite{Yue2024MMMU, Hudson2019GQA, Marino2019OK-VQA, Lu2022ScienceQA, liu2024MMBench}.
Many state-of-the-art VLMs have demonstrated high accuracy across these benchmarks, approaching human-level performance.

However, several studies have reported that VLMs do not necessarily demonstrate a rigorous understanding of images and texts~\cite{Thrush2022Winoground, Hall2023VisoGender}.
In particular, while VLMs are proficient at recognizing objects and actions, \textit{object state recognition} remains challenging~\cite{Xue2024HowToChange, Tateno2025MOST}.
Unlike object detection tasks, object state recognition inherently involves subtle distinctions, as a given object can exhibit a wide variety of states with varying degrees.
For instance, a refrigerator door described as ``open'' might be widely ajar or only slightly so, and when cooking, the difference between ingredients being roughly chopped versus finely diced represents a significant gradation in state.
Understanding differences between these subtle variations in object states is indeed required in numerous multimedia applications, such as developing robots for cooking meals or supporting vision-impaired persons with machine vision.
However, the lack of rigorous evaluation benchmarks for object state recognition prevents researchers from systematically identifying limitations in state-of-the-art VLMs.

In this paper, we address these limitations by introducing the \textbf{STAte and Transition UnderStanding Benchmark (STATUS Bench)},
the first benchmark for rigorously evaluating the ability of VLMs to understand object states.
First, we construct a high-quality human-annotated dataset of STATUS Bench, consisting of 404 quintuplets, each involving a pair of images, their corresponding object state descriptions, and a description of the state change, as shown in Figure~\ref{fig:top} (top).
We then introduce a novel evaluation scheme derived from the graph structure of these quintuplets. Specifically, as shown in Figure~\ref{fig:top} (bottom), we derive three evaluation tasks from the graph structure: object state identification (OSI), image retrieval (IR), and state change identification (SCI).
By requiring VLMs to perform these three tasks simultaneously, STATUS Bench enables a rigorous consistency evaluation, \textit{i.e.}, verifying whether VLMs genuinely understand subtle variations in object states.
Additionally, we introduce STATUS Train, a large-scale training dataset that serves as the largest resource to facilitate further research in this area.
This dataset is semi-automatically created using GPT-4o with manual filtering and contains 13 million descriptions of object states and state changes.

In our experiments, we extensively evaluate eight state-of-the-art VLMs.
We find that many open-weight VLMs exhibit chance-level zero-shot performance under the proposed rigorous consistency evaluation scheme.
For example, even when models perform well on the OSI task, there are often cases where they fail on the IR or SCI task for the same image pair, indicating inconsistencies in understanding object states.
To the best of our knowledge, we are the first to reveal and systematically analyze such phenomena in VLMs, providing insights into their limitations and suggesting directions for future improvement.
We also demonstrate that Qwen2.5-VL, which has the best zero-shot performance, achieves results comparable to Gemini 2.0 Flash after fine-tuning on STATUS Train.

In summary, our contributions are three-fold:

\noindent \textbf{1) STATUS Bench.}
We introduce a new benchmark for object state recognition with a novel evaluation scheme.
STATUS Bench enables rigorous consistency evaluation through OSI, IR and SCI tasks defined over quintuplet data (Figure~\ref{fig:top}).

\noindent \textbf{2) STATUS Train.}
We introduce the largest training dateset to date for object state recognition. STATUS Train consists of 13 million descriptions, more than 40 times larger than conventional datasets like CLEVR Change \cite{Park2019Clevr-change} and CLEVR MultiChange \cite{Qiu2021MultiClevrChange}.

\noindent \textbf{3) Extensive evaluations.}
We conduct experiments evaluating eight state-of-the-art VLMs. We first show that, in terms of rigorous consistency, most open-weight VLMs exhibit chance-level zero-shot performance. We then demonstrate that fine-tuning on STATUS Train is an effective first step toward addressing this limitation.

%% file: sections/02_related_work.tex
\begin{figure*}[t]
\centering
\includegraphics[width=\linewidth]{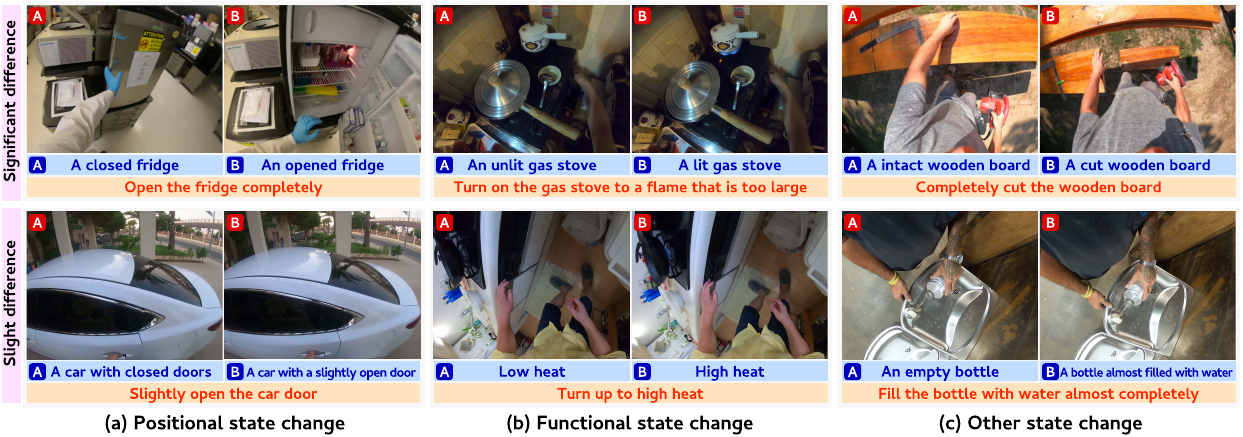}
\vspace{-20pt}
\caption{
Annotated image examples from STATUS Bench.
Each example is a quintuplet consisting of two images $\bm{(x^{a}, x^{b})}$, corresponding object state descriptions $\bm{(t^{a}, t^{b})}$, and a state change description $\bm{y}$.
}
\vspace{-10pt}
\label{fig:examples}
\end{figure*}

\newcommand{\xmark}{\ding{55}}
\begin{table}[t]
\setlength{\tabcolsep}{1pt}
\small
\centering
\caption{Statistical comparison of conventional change detection and captioning datasets.}
\vspace{-8pt}
\begin{tabular}{lcccc}
\toprule
\textbf{Dataset} & \textbf{Real} & \textbf{\# Image Pairs} & \textbf{\# Descriptions} & \textbf{\# Vocab.} \\
\midrule
VL-CMU-CD \cite{pablo2016VL-CMU-CD} & \checkmark & 1,362 & -- & -- \\
LEVIR-CD \cite{Chen2020LEVIR-CD} & \checkmark & 637 & -- & -- \\
PSCD \cite{sakurada2020PSCD} & \checkmark & 770 & -- & -- \\
SOCD \cite{doi2022SOCD} & \xmark & 15,000 & -- & -- \\
COCO Inpainted \cite{sachdeva2023changeyouwant} & \checkmark & 60,000 & -- & -- \\
Synthtext Change \cite{sachdeva2023changeyouwant} & \xmark & 5,000 & -- & -- \\
Kubric Change \cite{sachdeva2023changeyouwant} & \checkmark & 1,605 & -- & -- \\
VIRAT-STD \cite{sachdeva2023changeyouwant} & \checkmark & 1,000 & -- & -- \\
EGY-BCD \cite{holail2023egy-bcd} & \checkmark & 6,091 & -- & -- \\
ChangeNet \cite{ji2024changenet} & \checkmark & 31,000 & -- & -- \\
\midrule
CLEVR Change \cite{Park2019Clevr-change} & \xmark & 79,606 & 294,720 & 56 \\
CLEVR MultiChange \cite{Qiu2021MultiClevrChange} & \xmark & 60,000 & 300,000 & 51 \\
Spot-the-Diff \cite{jhamtani2018spotthediff} & \checkmark & 13,192 & 23,093 & 2,296 \\
LEVIR-CC \cite{Liu2022LEVIR-CC} & \checkmark & 10,077 & 50,385 & 995 \\
STVchrono \cite{Sun2024STVChrono} & \checkmark & 19,400 &  19,400 & -- \\
\textbf{STATUS (Ours)} & \checkmark &
4,333,298 & 12,999,894 & 11,015 \\
\bottomrule
\end{tabular}
\label{tab:dataset-comparison}
\end{table}

\section{Related work}

\subsection{Change Detection and Captioning Datasets}

Change detection and change captioning are tasks aimed at identifying differences between two images. Change detection specifically focuses on localizing the regions where changes have occurred. One of the most widely used datasets for this task is LEVIR-CD \cite{Chen2020LEVIR-CD}, which consists of high-resolution aerial images collected from Google Earth. Similar to LEVIR-CD, many change detection datasets are composed of aerial imagery \cite{Benedek2009SZTAKI, Daudt2018OSCD, Bourdis2011AICD, shen2021s2looking, ji2024changenet, holail2023egy-bcd}. In contrast, VL-CMU-CD \cite{pablo2016VL-CMU-CD}, PSCD \cite{sakurada2020PSCD}, and SOCD \cite{doi2022SOCD} are datasets that emphasize street view images for change detection. 
Sachdeva and Zisserman introduce four change detection datasets including synthetic and real-world scenes \cite{sachdeva2023changeyouwant}.
Change captioning, on the other hand, involves generating natural language descriptions of the differences between two images. Spot-the-Diff \cite{jhamtani2018spotthediff} and CLEVR-Change \cite{Park2019Clevr-change} are the most commonly used datasets for change captioning. Spot-the-Diff utilizes frames from the VIRAT dataset \cite{oh2011virat} for change captioning tasks at specific locations, while CLEVR-Change is a synthetic image dataset generated using the CLEVR engine \cite{Johnson2017CLEVR}. CLEVR-Multi-Change \cite{Qiu2021MultiClevrChange} extends CLEVR-Change by increasing the number of change differences. Additionally, datasets such as LEVIR-CC \cite{Liu2022LEVIR-CC} and STVChrono \cite{Sun2024STVChrono} focus on recognizing longer-term changes.
However, these datasets primarily focus on changes related to object existence, position, or color, which causes challenges when applying models to recognize changes in object states. In this study, we aim to generate captions that describe differences between two images depicting objects in different states.

\subsection{Object State Recognition}

Object state recognition involves identifying or localizing the state of an object, as well as detecting changes in that state. Pioneering dataset in this field is MIT-States \cite{Isola2015MIT-States}, which focuses on object states. This dataset provides images annotated with specific object–state combinations (e.g., sliced apple or empty box) and is primarily designed to facilitate recognition of object-associated states.
The cooking domain occupies an important portion in object change recognition~\cite{stein2013combing,zhou2018Youcook2,Damen2018EPICKITCHENS,shirai2022visual,Saini2023ChopNLearn,soucek2024multitask}.
Joint manipulation action and object state recognition are performed in about 20,000 annotated tracks in third person vision \cite{Alayrac_2017_ICCV}.
ChangeIt \cite{soucek2022changeit}, COIN \cite{tang2019coin} and HowToChange \cite{Xue2024HowToChange} focus on the task of localizing the time spans of some actions from the initial state to the end state.
OSDD \cite{gouidis2022OSDD} aimed to provide annotations for objects situated in real-world backgrounds.
MOST \cite{Tateno2025MOST} emphasizes the localization of object states in videos, including transitions between states. OSCaR \cite{Nguyen2024OSCaR} is a dataset dedicated to generating explanations for object states in open-domain settings. Additionally, multi-task datasets such as MMVP \cite{Tong2024MMVP} include object state recognition as one of several subtasks.
Despite these efforts, existing datasets lack a large number of training instances necessary for nuanced and diverse recognition of object states. Our objective is to go beyond merely recognizing object states in limited scenarios and to capture how these states are realized across a wide range of situations.

\subsection{Vision-Language Models}
Recent research in the field of Large Language Models (LLMs) has led to advancements in their ability to understand visual information. Among the open-source VLMs, LLaVA \cite{liu2023LLaVA} is one of the most widely recognized.
Other notable models in this domain include PaLI \cite{chen2023pali} and InternVL \cite{Chen2024InternVL}.
Additionally, several studies have introduced LLMs capable of processing multiple image inputs. Prominent examples of such models include the VILA series \cite{Lin2024VILA, liu2024nvila}, the Qwen series \cite{bai2023qwen, wang2024qwen2vl, bai2025qwen25vl}, and the Video LLaMA series \cite{cheng2024videollama2}. These models are designed to handle multiple image inputs and demonstrate significant advancements in this area. Moreover, models like LLaVA OneVision \cite{li2024llavaonevision} and Intern2.5VL \cite{chen2025intern25vl} extend the capabilities of their predecessors, enabling them to perform inference on multiple image inputs.
In addition to open-source VLMs, several high-performing API models, such as GPT and Gemini, also offer the ability to recognize multiple image inputs. The latest models of these series, GPT-4o \cite{OpenAI2024GPT-4o} and Gemini 2.0 Flash \cite{geminiteam2024gemini}, have demonstrated strong performance across various vision and language tasks, further pushing the boundaries of what is achievable in this domain.

Their abilities are measured using benchmarks focused on multiple image inference \cite{suhr2019nlvr2, Yue2024MMMU}. Similarly, we evaluate them in the context of multiple image inference for object state recognition.

%% file: sections/03_status.tex
\begin{figure*}
\centering
\includegraphics[width=\linewidth]{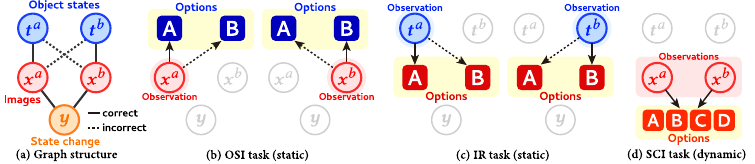}
\vspace{-16pt}
\caption{
Proposed rigorous consistency evaluation scheme over quintuplet data.
(a) Quintuplet consists of an image pair $(\bm{x}^{a}, \bm{x}^{b})$, a state description pair $(\bm{t}^{a}, \bm{t}^{b})$, and a state change description $\bm{y}$. Edges represent dependency, where dashed lines indicate incorrect links.
(b) OSI task assumes that a single image is observed. VLMs are tasked with selecting the correct description.
(c) IR task assumes that a single state description is observed. VLMs are tasked with selecting the correct image.
(d) SCI task assumes that two images are observed.
VLMs are tasked with selecting the correct stat change description.
}
\vspace{-8pt}
\label{fig:schemes}
\end{figure*}

\section{STATUS Bench}

We introduce the \textbf{STATUS Bench}, the first benchmark for rigorously evaluating the ability of VLMs to understand subtle variations in object states.
STATUS Bench consists of two evaluation scenarios for consistent evaluation: static and dynamic.
The static scenario encompasses the OSI and IR tasks, both of which focus on object states within individual images, such as the amount of water in a bucket or the degree to which a door is open.
In contrast, the dynamic scenario involves the SCI task, which targets state changes by providing VLMs with pairs of images showing transitions in object states.
Combining these two scenarios, we propose \textbf{Rigorous Overall Accuracy (ROA)}, a novel metric to measure the rigorous consistency of VLMs in recognizing object states.
The ROA metric emphasizes rigor and consistency by evaluating performance across all three tasks (OSI, IR, and SCI) simultaneously on each image pair.
Notably, since current state-of-the-art VLMs still significantly struggle with capturing such subtle object state distinctions, STATUS Bench is entirely hand-curated by expert annotators, ensuring a high-quality evaluation of model performance.

\subsection{Dataset Structure}

To provide a foundation for the rigorous consistency evaluation, we first construct a dataset consisting of image-text quintuplets 
$\mathcal{D} = \{(\bm{x}^{a}_{i}, \bm{x}^{b}_{i}, \bm{t}^{a}_{i}, \bm{t}^{b}_{i}, \bm{y}_{i})\}_{i=1}^{N}$.
Here, $(\bm{x}^{a}_{i}, \bm{x}^{b}_{i})$ is a pair of images differing slightly only in object states;
$\bm{t}^{a}_{i}$ and $\bm{t}^{b}_{i}$ are textual descriptions of the object states for $\bm{x}^{a}_{i}$ and $\bm{x}^{b}_{i}$, respectively;
and $\bm{y}_{i}$ is a textual description of the state change from $\bm{x}^{a}_{i}$ to $\bm{x}^{b}_{i}$.
The dataset involves 404 high-quality quintuplets that are carefully chosen and manually annotated.
As shown in Figure~\ref{fig:examples}, the paired images are often visually similar to each other but differ in terms of the object state.
Each state description is composed of a noun indicating the object and its state, while the state change description uses a verb to describe the change in state.

\myparagraph{Source images}
We selected Ego4D~\cite{Grauman2022Ego4D} as a source to extract image pairs due to its extensive scale and rich diversity of real-world contexts.
We manually selected key frame pairs from which object state transitions can be clearly identified.
The key frame extraction procedure consists of the following three steps.
First, candidate video clips are pre-filtered to ensure visual clarity in terms of high resolution and minimal blur. Specifically, we apply blur detection based on Laplacian variance using OpenCV to videos with resolutions higher than 540p and select 200 minimal blur videos.
Second, we visually examined each video segment to identify clear instances of object state transitions.
Finally, 404 pairs of key frames that clearly illustrate object state change are manually extracted.

\myparagraph{State descriptions}
For each key frame image, we annotate a concise textual description of the object name and its state.
Specifically, for each key frame pair $(\bm{x}^{a}_{i}, \bm{x}^{b}_{i})$, expert annotators identify the primary object in focus and specify subtle variations in its state, resulting in a pair of state descriptions $(\bm{t}^{a}_{i}, \bm{t}^{b}_{i})$, \textit{e.g.}, ``a slightly opened door'' and ``a fully opened door.''
Each description strictly adheres to a structured format, combining an object noun with a descriptive adjective or phrase indicating the state.

\myparagraph{State change descriptions}
We also annotate a textual description $\bm{y}_{i}$ capturing the transition in object state for each key frame pair $(\bm{x}^{a}_{i}, \bm{x}^{b}_{i})$.
Annotators use descriptive verbs to indicate the nature of the state change, \textit{e.g.}, ``open the door'' and ``turn on the light.''

\myparagraph{Graphical model}
Figure~\ref{fig:schemes}(a) shows the graph structure of the quintuplet, with solid and dashed lines representing correct and incorrect dependencies, respectively.
In the proposed evaluation scheme, we prompt VLMs to predict each element of the quintuplet across static and dynamic scenarios.

\subsection{Static Scenario}

The static scenario evaluates the model's performance to identify object states within individual images through the OSI and IR tasks.
In the OSI task, VLMs are required to predict the object state $\bm{t}$ given an input image $\bm{x}$.
Conversely, in the IR task, VLMs are tasked with retrieving the correct image $\bm{x}$ given an object state $\bm{t}$.
These two tasks are defined over the graph structure as follows.

\myparagraph{OSI Task}
Given a quintuplet $(\bm{x}^{a}_{i}, \bm{x}^{b}_{i}, \bm{t}^{a}_{i}, \bm{t}^{b}_{i}, \bm{y}_{i}) \in \mathcal{D}$, the OSI task assumes that each of the two images ($\bm{x}^{a}_{i}$ and $\bm{x}^{b}_{i}$) is observed separately, and prompts VLMs to predict the correct object state from two textual options $\bm{T} = \{\bm{t}^{a}_{i}, \bm{t}^{b}_{i}\}$ for each image individually.
Specifically, this procedure is formulated as
\begin{align}
\hat{\bm{t}}_{i}^{a} = F(\bm{p}_{\hspace{0.3pt}\text{\tiny \textbf{OSI}}}, \bm{T}, \bm{x}^{a}_{i}),\quad
\hat{\bm{t}}_{i}^{b} = F(\bm{p}_{\hspace{0.3pt}\text{\tiny \textbf{OSI}}}, \bm{T}, \bm{x}^{b}_{i}),
\end{align}
where $F$ is a VLM,
$\bm{p}_{\hspace{0.3pt}\text{\tiny \textbf{OSI}}}$ is the instruction prompt,
and
$\hat{\bm{t}}_{i}^{a}, \hat{\bm{t}}_{i}^{b}$ are predicted object states for the two input images $\bm{x}^{a}_{i}$ and $\hat{\bm{x}}_{i}^{b}$, respectively.
Here, the inputs are concatenated at the token level.
Then, the accuracy for this task is computed as
\begin{align}
\mathrm{Acc}_{\hspace{1pt}\text{\footnotesize OSI}}
=
\frac{1}{2N}
\sum_{i=1}^{N}
(\llbracket \hat{\bm{t}}_{i}^{a} = \bm{t}^{a}_{i} \rrbracket + \llbracket \hat{\bm{t}}_{i}^{b} = \bm{t}^{b}_{i} \rrbracket),
\end{align}
where $\llbracket P \rrbracket$ takes the value $1$ if the proposition $P$ is true and $0$ otherwise.
This accuracy measures the fraction of descriptions for which the VLM's predictions exactly match the ground truth.
In the instruction prompt, we label $\bm{t}^{a}_{i}$ and $\bm{t}^{b}_{i}$ with ``A'' and ``B,'' respectively, and instruct the VLM to select the appropriate one.
The graphical model is shown in Figure~\ref{fig:schemes}(b), where $\bm{t}^{a}_{i}$ and $\bm{t}^{b}_{i}$ correspond to options ``A'' and ``B'', respectively.

\myparagraph{IR Task}
This task requires the VLM to retrieve the correct image given an object state description.
Specifically, each state description serves as a query, and the VLM selects the appropriate image from two options $\bm{X} = \{\bm{x}^{a}_{i}, \bm{x}^{b}_{i}\}$.
This procedure is formulated as
\begin{align}
\hat{\bm{x}}_{i}^{a} = F(\bm{p}_{\hspace{0.3pt}\text{\tiny \textbf{IR}}}, \bm{t}^{a}_{i}, \bm{X}),\quad
\hat{\bm{x}}_{i}^{b} = F(\bm{p}_{\hspace{0.3pt}\text{\tiny \textbf{IR}}}, \bm{t}^{b}_{i}, \bm{X}),
\end{align}
where $\hat{\bm{x}}_{i}^{a}, \hat{\bm{x}}_{i}^{b}$ are images selected by the VLM, and $\bm{p}_{\hspace{0.3pt}\text{\tiny \textbf{IR}}}$ is the instruction prompt.
Here, we label two images with ``A'' and ``B,'' and prompt the VLM to output the index of the selected image.
As shown in Figure~\ref{fig:schemes}(c), the graphical model is complementary to that of the OSI task, providing a means to measure consistency.
Similar to OSI, the accuracy is computed by
\begin{align}
\mathrm{Acc}_{\hspace{1pt}\text{\footnotesize IR}}
=
\frac{1}{2N}
\sum_{i=1}^{N}
(\llbracket \hat{\bm{x}}_{i}^{a} = \bm{x}^{a}_{i} \rrbracket + \llbracket \hat{\bm{x}}_{i}^{b} = \bm{x}^{b}_{i} \rrbracket).
\end{align}

\subsection{Dynamic Scenario}

The dynamic scenario evaluates the model's ability to identify object state changes.
This scenario relies on the SCI task, in which VLMs must predict the state change given a pair of images.

\myparagraph{SCI Task}
As shown in Figure~\ref{fig:schemes}(d), this task assumes that paired images are given. The VLM predicts the state change as
\begin{align}
\hat{\bm{y}}_{i} = F(\bm{p}_{\hspace{0.3pt}\text{\tiny \textbf{SCI}}}, \bm{X}), \end{align}
where $\bm{p}_{\hspace{0.3pt}\text{\tiny \textbf{SCI}}}$ is an instruction prompt and $\bm{X} = \{\bm{x}^{a}_{i}, \bm{x}^{b}_{i}\}$ is a pair of images.
In the instruction prompt, we provide manually crafted two options consisting of the ground truth description and another misleading description.
Then, the accuracy is computed as
\begin{align}
\mathrm{Acc}_{\hspace{1pt}\text{\footnotesize SCI}}
=
\frac{1}{N} \sum_{i=1}^{N} \llbracket \hat{\bm{y}}_{i} = \bm{y}_{i} \rrbracket.
\end{align}

\subsection{Rigorous Consistency Evaluation}

To rigorously measure consistency, we evaluate whether the VLM can simultaneously perform the three tasks (OSI, IR, and SCI).
To this end, we introduce the rigorous accuracy metrics.

\myparagraph{Rigorous Accuracy}
For the OSI and IR tasks, Rigorous Accuracy (RAcc) measures the fraction of quintuplets for which the VLM's outputs exactly match the ground-truth.
Specifically, we define RAcc as follows:
\begin{align}
\mathrm{RAcc}_{\hspace{1pt}\text{\footnotesize OSI}}
&=
\frac{1}{N}
\sum_{i=1}^{N}
\llbracket (\hat{\bm{t}}_{i}^{a} = \bm{t}^{a}_{i}) \land (\hat{\bm{t}}_{i}^{b} = \bm{t}^{b}_{i}) \rrbracket,\\
\mathrm{RAcc}_{\hspace{1pt}\text{\footnotesize IR}}
&=
\frac{1}{N}
\sum_{i=1}^{N}
\llbracket (\hat{\bm{x}}_{i}^{a} = \bm{x}^{a}_{i}) \land (\hat{\bm{x}}_{i}^{b} = \bm{x}^{b}_{i}) \rrbracket,
\end{align}
where $\land$ indicates the logical AND operation.
By requiring simultaneous correctness for the two predictions in each task, these metrics enhance the rigor of evaluation.
For the SCI task, we increase the number of misleading options to three and refer to this accuracy as RAcc.
All of these rigorous accuracies have a chance rate of 25\%.

\myparagraph{Rigorous Overall Accuracy}
Finally, we define ROA by extending RAcc to cover all tasks:
\begin{align}
\label{eq:roa}
\mathrm{ROA}
=
\frac{1}{N}
\sum_{i=1}^{N}
\llbracket
&(\hat{\bm{x}}_{i}^{a} = \bm{x}^{a}_{i}) \land (\hat{\bm{x}}_{i}^{b} = \bm{x}^{b}_{i}) \land \nonumber \\[-8pt]
&(\hat{\bm{t}}_{i}^{a} = \bm{t}^{a}_{i}) \land (\hat{\bm{t}}_{i}^{b} = \bm{t}^{b}_{i}) \land
(\hat{\bm{y}}_{i} = \bm{y}_{i})
\rrbracket.
\end{align}
This metric requires simultaneous correctness across all five elements of the quintuple, thereby enabling more consistency evaluation.
We also define the standard overall accuracy (OA) by mean of the three accuracies:
\begin{align}
\mathrm{OA}
=
\frac{1}{3} (
\mathrm{Acc}_{\hspace{1pt}\text{\footnotesize OSI}}
+
\mathrm{Acc}_{\hspace{1pt}\text{\footnotesize IR}}
+
\mathrm{Acc}_{\hspace{1pt}\text{\footnotesize SCI}}
)
\end{align}
Although high ROA scores imply high OA scores, its reverse does not hold.
The chance-level performance for ROA is $0.5^{4} \times 0.25 \simeq 1.56\%$.
Our experiments reveal that many VLMs perform well in terms of OA but achieve ROA performance at or even below this chance level.

\begin{figure}
\centering
\includegraphics[width=\linewidth]{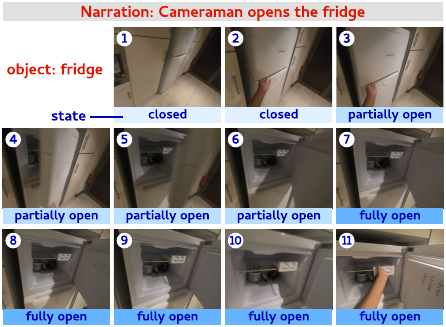}
\vspace{-20pt}
\caption{STATUS Train is a  dataset semi-automatically created from video and narration data of Ego4D. Eleven key frames are extracted for each narrated event and used to generate time-consistent object state descriptions.}
\label{fig:fridge}
\vspace{-10pt}
\end{figure}

\definecolor{Green}{RGB}{0,200,0}
\definecolor{Red}{RGB}{230,0,0}
\begin{table*}[t]
\centering
\setlength{\tabcolsep}{7.2pt}
\caption{Comparison of eight VLMs on STATUS Bench. For Overall Accuracy (OA) and Rigorous Overall Accuracy (ROA), the differences from the chance rate are indicated in red or green.}
\vspace{-8pt}
\label{tab:status_bench}
\begin{tabular}{l|cccc|cccc}
\toprule
\multirow{2}{*}{\textbf{Model}} & \multicolumn{4}{c|}{\textbf{Standard Accuracy}} & \multicolumn{4}{c}{\textbf{Rigorous Accuracy}} \\
& $\bm{\mathrm{Acc}_{\mathrm{OSI}}}$ &
$\bm{\mathrm{Acc}_{\mathrm{IR}}}$ & 
$\bm{\mathrm{Acc}_{\mathrm{SCI}}}$ & $\bm{\mathrm{OA}}$
& $\bm{\mathrm{RAcc}_{\mathrm{OSI}}}$ &
$\bm{\mathrm{RAcc}_{\mathrm{IR}}}$ & 
$\bm{\mathrm{RAcc}_{\mathrm{SCI}}}$ & $\bm{\mathrm{ROA}}$
\\
\midrule
Random & 50.00 & 50.00 & 50.00 & 50.00 & 25.00 & 25.00 & 25.00 & 1.56 \\
\midrule
NVILA~\cite{liu2024nvila} & 59.03 & 50.25 & 51.98 & 53.75 \textcolor{Green}{~(+3.75)} & 22.52 & 0.74 & 27.23 & 0.25 \textcolor{Red}{~(-1.31)} \\
Llama-3.2-Vision-Inst~\cite{grattafiori2024llama3} & 56.56 & 49.26 & 45.54 & 50.45 \textcolor{Green}{~(+0.45)} & 25.74 & 7.92 & 27.48 & 0.25 \textcolor{Red}{~(-1.31)} \\
VideoLLaMA3~\cite{cheng2024videollama2} & 54.21 & 52.97 & 45.79 & 50.99 \textcolor{Green}{~(+0.99)} & 21.04 & 20.79 & 18.81 & 0.74 \textcolor{Red}{~(-0.82)} \\
Intern2.5VL~\cite{chen2025intern25vl} & 56.56 & 47.77 & 54.46 & 52.93 \textcolor{Green}{~(+2.93)} & 30.45 & 24.26 & 20.79 & 1.24 \textcolor{Red}{~(-0.32)} \\
LLaVA-OneVision~\cite{li2024llavaonevision} & 63.86 & 56.19 & 53.22 & 57.76 \textcolor{Green}{~(+7.76)} & 31.19 & 15.10 & 30.94 & 1.98 \textcolor{Green}{~(+0.42)} \\
\rowcolor[HTML]{ffede6} LLaVA-OneVision (fine-tuned) & \textbf{63.99} & 56.31 & \textbf{57.18} & 59.16 \textcolor{Green}{~(+9.16)} & 31.19 & 15.35 & \textbf{38.12} & 2.72 \textcolor{Green}{~(+1.16)} \\
Qwen2.5VL-Inst~\cite{bai2025qwen25vl} & 46.53 & 60.89 & 53.47 & 53.63 \textcolor{Green}{~(+3.63)} & 18.32 & 32.67 & 33.42 & 5.45 \textcolor{Green}{~(+3.89)} \\
\rowcolor[HTML]{ffede6} Qwen2.5VL-Inst (fine-tuned) & 63.49 & \textbf{61.39} & 55.94 & \textbf{60.27} \textcolor{Green}{~(+10.27)} & \textbf{33.91} & \textbf{34.16} & 35.89 & \textbf{7.18} \textcolor{Green}{~(+5.62)} \\
\midrule
Gemini 2.0 Flash~\cite{geminiteam2024gemini} & 61.88 & 60.02 & 54.46 & 58.79 \textcolor{Green}{~(+8.79)} & 26.73 & 31.19 & 36.39 & 6.19 \textcolor{Green}{~(+4.63)} \\
GPT-4o~\cite{OpenAI2024GPT-4o} & \textbf{64.11} & \textbf{71.53} & \textbf{61.39} & \textbf{65.68} \textcolor{Green}{~(+15.68)} & \textbf{32.67} & \textbf{50.50} & \textbf{45.05} & \textbf{10.89} \textcolor{Green}{~(+9.33)} \\
\bottomrule
\end{tabular}
\end{table*}

\section{STATUS Train}

To alleviate the difficulties of capturing subtle variations in object states, we further introduce STATUS Train, a semi-automatically generated training dataset containing 4.3 million examples.

\myparagraph{Source images}
We leveraged the narrations provided in Ego4D, which include timestamps and descriptions of human actions, to extract key frames.
Specifically, from each video not used in STATUS Bench, we extracted 11 key frames spanning 2 seconds before and after each narrated event as shown in Figure~\ref{fig:fridge}.
We then performed syntactic parsing to filter out narrations that merely describe object movement or camera shifts, yielding 4,046 unique verbs.
Finally, we manually selected verbs that represent state changes, ensuring the dataset contained a higher proportion of true state changes.

\myparagraph{Object State Descriptions}
We employed GPT-4o \cite{OpenAI2024GPT-4o} to generate object state descriptions for each key frame.
To focus on variations in object states, we designed a prompt to include the state of the objects referenced in the narration, while avoiding unnecessary details about the background or surrounding elements.
We then fed all 11 key frames in chronological order with the prompt into GPT-4o.
This approach not only allows the model to leverage multiple comparisons when generating descriptions but also helps enforce temporal consistency, preventing physically contradictory captions.
For example, in a sequence labeled ``opening a cabinet'', the model would generate ``closed cabinet'' for early frames, ``partially open cabinet'' for midway frames, and ``fully open cabinet'' for late frames.

\myparagraph{State Change Descriptions}
Finally, we generated state change descriptions.
We selected 4.3 million pairs of key frame images with differing object-state descriptions and then generated state change descriptions based on the object state descriptions and images.
Although our dataset construction process relies on GPT-4o,
integrating detailed narration data and utilizing a higher number of key frames enabled the creation of a large-scale, high-quality dataset.

%% file: sections/04_experiments.tex
\section{Experiments}

We present extensive experiments on the STATUS Bench, evaluating eight state-of-the-art VLMs under the proposed rigorous evaluation scheme, where our primary aim is to assess not only accuracy but also model consistency.
Our results show that most open-weight VLMs exhibit chance-level zero-shot performance, indicating a lack of consistency when confronted with subtle changes in the input.
Through a detailed analysis, we identify that this inconsistency arises because current models tend to rely on superficial cues, rather than capturing the fine-grained visual differences. Specifically, although differences are often preserved in intermediate representations, a miscalibration in the final linear head frequently results in inconsistent predictions.
We also demonstrate that Qwen2.5-VL, once fine-tuned on STATUS Train, is able to achieve performance on par with Gemini 2.0 Flash.

\subsection{Experimental Settings}

\myparagraph{Models}
We evaluate eight state-of-the-art VLMs:
NVILA-8B \cite{liu2024nvila},
Llama-3.2-Vision-Instruct-11B~\cite{grattafiori2024llama3},
VideoLLaMA3-7B~\cite{cheng2024videollama2}, Intern2.5-VL-8B~\cite{chen2025intern25vl},
Qwen2.5-VL-Instruct-7B \cite{bai2025qwen25vl},
LLaVA-OneVision-7B \cite{li2024llavaonevision}, Gemini 2.0 Flash \cite{geminiteam2024gemini} and GPT-4o \cite{OpenAI2024GPT-4o}.
The first six are open-weight VLMs, while the remaining two are API-based.
We selected these VLMs due to their capability to handle multiple images within a single inference step.

\myparagraph{Datasets and metrics}
STATUS Bench is used to measure the performance of VLMs in terms of the proposed ROA metric in Eq.~\eqref{eq:roa}.
We also report standard and rigorous accuracies for each task.
STATUS Train is used for fine-tuning, where LoRA and Adam are used with cross-entropy loss for one epoch with a learning rate of $10^{-6}$.

\subsection{Experimental Results}

\myparagraph{Benchmark Results}
Table~\ref{tab:status_bench} summarizes the evaluation results on STATUS Bench.
As shown, recognizing subtle variations in object states remains a significant challenge for most VLMs.
Among the six open-weight VLMs, Qwen2.5VL-Inst achieves the highest performance; however, it still produces consistent predictions only 5.45\%. The other VLMs remain close to chance-level performance.
From a graphical modeling perspective, if the prediction is correct within one of OSI, IR, or SCI task, consistency should help determine the remaining solutions.
In other words, with perfect consistency, the standard accuracy and rigorous accuracy would coincide.
However, across all evaluated VLMs, we observe a substantial drop in rigorous accuracy compared to standard accuracy.
These findings underscore that even state-of-the-art VLMs exhibit limited consistency in their comprehension of object states.

\begin{table*}[t]
\centering
\setlength{\tabcolsep}{6pt}
\caption{Performance by state categories. Rigorous accuracy is used as the evaluation metric.}
\vspace{-8pt}
\label{tab:state_category}
\begin{tabular}{l|cccc|cccc|cccc}
\toprule
\multirow{2}{*}{Model} & \multicolumn{4}{c|}{\textbf{Positional states}} & \multicolumn{4}{c|}{\textbf{Functional states}} & \multicolumn{4}{c}{\textbf{Others}} \\ 
& 
OSI & IR & SCI & ROA &
OSI & IR & SCI & ROA &
OSI & IR & SCI & ROA \\
\midrule
Random & 25.00 & 25.00 & 25.00 & 1.56 & 25.00 & 25.00 & 25.00 & 1.56 & 25.00 & 25.00 & 25.00 & 1.56 \\
\midrule
NVILA & 30.43 & 0.62 & 26.71 & 0.00 & 20.56 & 1.87 & 33.64 & 0.93 & 14.71 & 0.00 & 22.79 & 0.00 \\
Llama-3.2-Vision-Inst & 23.60 & 9.32 & 31.06 & 0.00 & 28.04 & 8.41 & 29.91 & 0.93 & 26.47 & 5.88 & 21.32 & 0.00 \\
VideoLLaMA3 & 28.57 & 22.98 & 10.56 & 0.62 & 16.82 & 21.50 & 14.02 & 0.93 & 15.44 & 17.65 & 11.03 & 0.74 \\
Intern2.5VL & 34.16 & 25.47 & 19.88 & 1.24 & 32.71 & 23.36 & 21.50 & 1.87 & 24.26 & 23.53 & 21.32 & 0.74 \\
LLaVA-OneVision & 40.49 & 20.24 & 31.17 & 2.83 & 24.30 & 13.08 & 37.38 & 1.87 & 27.21 & 10.29 & 27.94 & 0.74 \\
\rowcolor[HTML]{ffede6} LLaVA-OneVision (fine-tuned) & 39.75 & 21.12 & 37.89 & 4.35 & 25.23 & 13.08 & \textbf{42.06} & 2.80 & 25.74 & 10.29 & \textbf{35.29} & 0.74 \\
Qwen2.5VL-Inst & 26.71 & 35.40 & \textbf{42.24} & 10.56 & 13.08 & \textbf{33.64} & 28.97 & 1.87 & 12.50 & \textbf{28.68} & 26.47 & 2.21 \\
\rowcolor[HTML]{ffede6} Qwen2.5VL-Inst (fine-tuned) & \textbf{41.61} & \textbf{44.10} & 40.99 & \textbf{11.18} & \textbf{28.04} & 31.78 & 37.38 & \textbf{3.74} & \textbf{29.41} & 24.26 & 28.68 & \textbf{5.15} \\
\midrule
Gemini 2.0 Flash & 29.81 & 39.75 & 42.86 & 9.32 & 25.23 & 25.23 & 33.64 & 4.67 & 24.26 & 25.00 & 30.15 & 3.68 \\
GPT-4o & \textbf{38.51} & \textbf{57.14} & \textbf{48.45} & \textbf{17.39} & \textbf{29.91} & \textbf{48.60} & \textbf{44.86} & \textbf{5.61} & \textbf{27.94} & \textbf{44.12} & \textbf{41.18} & \textbf{7.35} \\
\bottomrule
\end{tabular}
\end{table*}

\begin{table}[t]
\centering
\setlength{\tabcolsep}{1.5pt}
\caption{Performance by image difference levels. Rigorous accuracy is used as the evaluation metric. $^{\dagger}$ indicates fine-tuned models.}
\vspace{-8pt}
\label{tab:difference_level}
\begin{tabular}{l|cccc|cccc}
\toprule
\multirow{2}{*}{Model} & \multicolumn{4}{c|}{\textbf{Slight}} & \multicolumn{4}{c}{\textbf{Significant}} \\
& OSI & IR & SR & ROA & OSI & IR & SR & ROA \\
\midrule
Random & 25.00 & 25.00 & 25.00 & 1.56 & 25.00 & 25.00 & 25.00 & 1.56 \\
\midrule
NVILA & 9.55 & 0.00 & 25.48 & 0.00 & 30.77 & 1.21 & 28.34 & 0.40 \\
Llama-3.2-Vision & 17.20 & 7.01 & 26.11 & 0.00 & 31.17 & 8.50 & 28.34 & 0.40 \\
VideoLLaMA3 & 15.92 & 14.65 & 9.55 & 0.00 & 24.29 & 24.70 & 12.96 & 1.21 \\
Intern2.5VL & 25.48 & 25.48 & 18.47 & 0.64 & 33.60 & 23.48 & 22.27 & 1.62 \\
LLaVA-OneVision & 16.56 & 5.73 & 30.57 & 0.00 & 40.49 & 20.24 & 31.17 & 2.83 \\
\rowcolor[HTML]{ffede6} LLaVA-OneVision$^{\dagger}$ & 17.20 & 7.01 & 35.67 & 1.27 & 40.08 & 20.65 & 39.68 & 3.64 \\
Qwen2.5VL-Inst. & 9.55 & 19.11 & 26.11 & 1.91 & 23.89 & 41.30 & 38.06 & 7.69 \\
\rowcolor[HTML]{ffede6} Qwen2.5VL-Inst.$^{\dagger}$ & 19.75 & 17.20 & 29.30 & 2.55 & 42.91 & 44.94 & 40.08 & 10.12 \\
\midrule
Gemini & 19.75 & 21.02 & 28.66 & 0.64 & 31.17 & 37.25 & 40.89 & 9.72 \\
GPT-4o & 22.29 & 29.94 & 40.76 & 3.18 & 39.27 & 63.56 & 47.77 & 15.79 \\
\bottomrule
\end{tabular}
\vspace{-12pt}
\end{table}

\myparagraph{Comparison with GPT-4o and Gemini}
Table~\ref{tab:status_bench} also reports results for two API-based VLMs, namely OpenAI GPT-4o and Google Gemini. We observe that these two models outperform the open-weight VLMs, indicating that training on larger and more diverse data can further enhance the rigorous consistency required by STATUS Bench.

\myparagraph{Finetuning results}
We further train the best performing models with the STATUS Train set. For this purpose, we use LoRA considering the computational cost and the amount of STATUS Train.
The results with model training are highlighted in
Table~\ref{tab:status_bench}.
Qwen2.5VL-Inst achieves competitive or surpassing performance with Gemini,
demonstrating that STATUS Train is a valuable resource for improving performances of object state recognition.

\subsection{Analysis}

\myparagraph{Is Model Performance Sensitive to State Categories?}
To analyze whether model performance varies across different types of object states, we subdivide state annotations into three categories: \textit{Positional}, \textit{Functional}, and \textit{Others}.
Positional states capture spatial arrangements or orientations (Figure~\ref{fig:examples}(a)). 
Functional states reflect the property of an object (Figure~\ref{fig:examples}(b)). 
All remaining states that do not neatly fit into these two categories are grouped under Others (Figure~\ref{fig:examples}(c)).
Table~\ref{tab:state_category} presents a detailed performance breakdown across these categories.
We observe that Positional states appear comparatively easier, with Qwen2.5VL-Inst showing significantly better performance in terms of ROA.
This suggests that physical relationships may align better with the pretraining objectives of current VLMs.
In contrast, Functional and Others pose greater challenges, reflecting limitations in the models' abilities to capture abstract or contextual state properties.
Although fine-tuning leads to performance improvements across all categories, consistently recognizing non-positional states remains particularly difficult, underscoring the need for further research into enhancing semantic understanding and logical consistency.

\begin{figure*}
\centering
\includegraphics[width=0.95\linewidth]{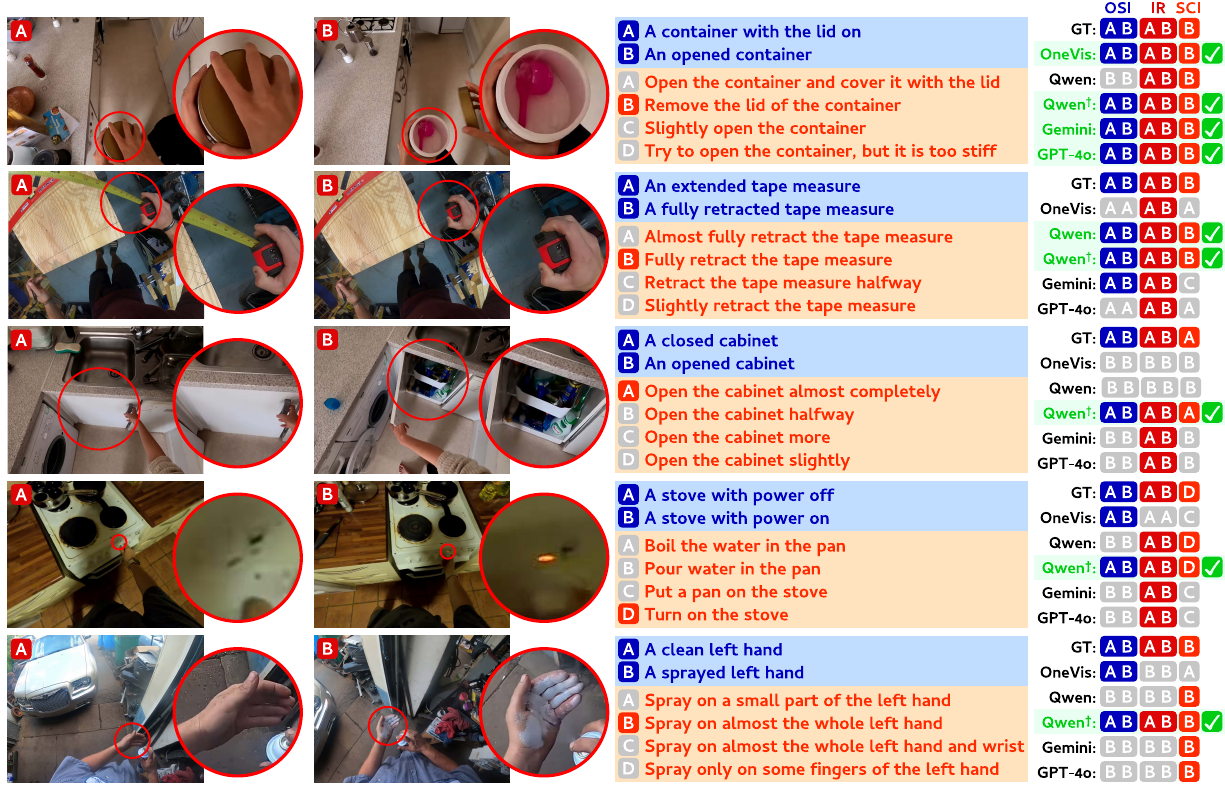}
\vspace{-10pt}
\caption{Example results. Paired predictions for OSI and IR, and a single prediction for SCI are shown for each model. Consistently correct predictions that contribute to higher ROA values are marked in green. $^{\dagger}$ indicates a fine-tuned model.}
\label{fig:error_analysis}
\end{figure*}

\begin{figure}
\centering
\includegraphics[width=0.90\linewidth]{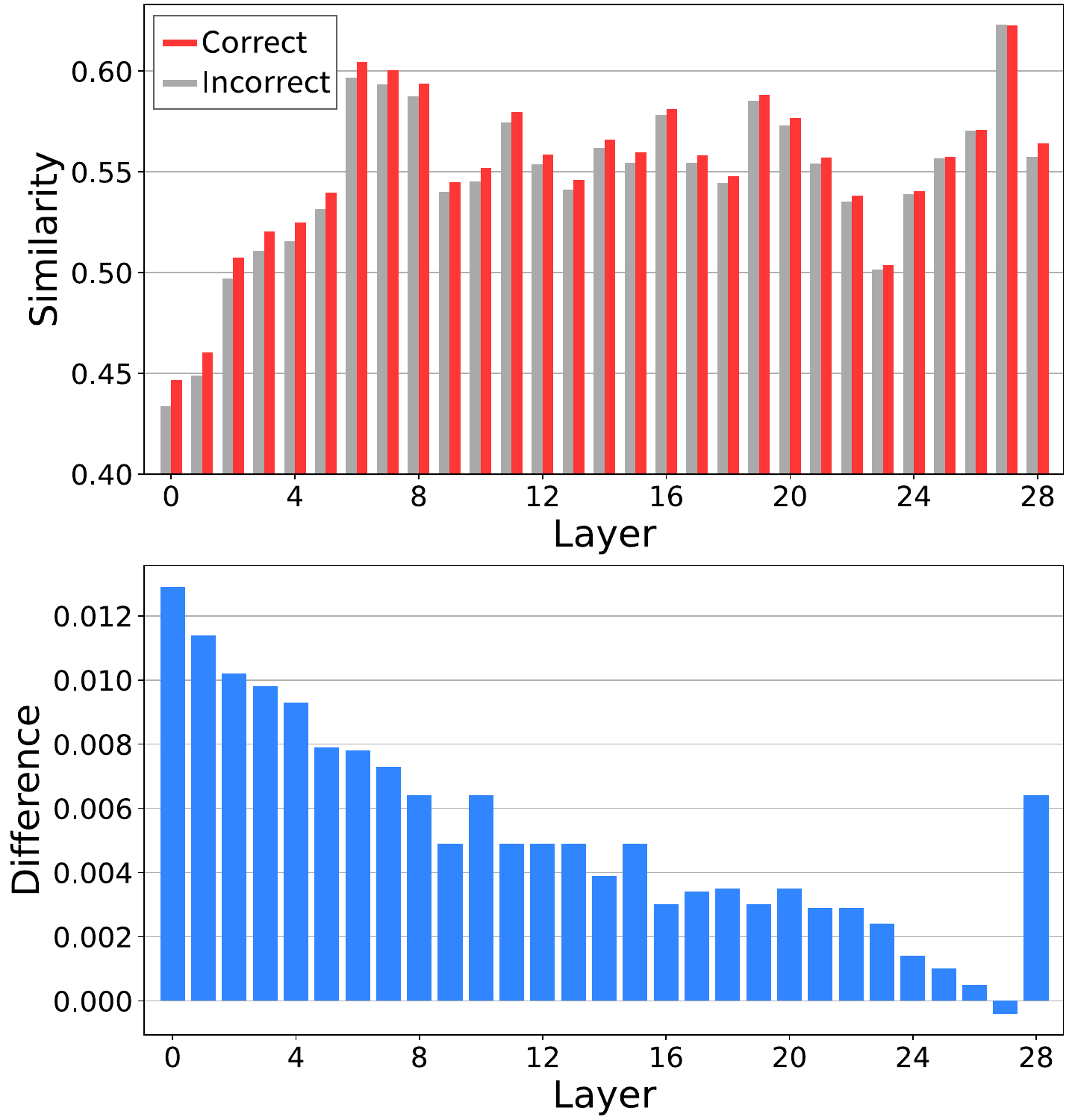}
\vspace{-8pt}
\caption{Analysis of hidden feature similarity. Top: Average cosine similarity per layer. Bottom: Difference between similarities.}
\label{fig:similarity}
\vspace{-14pt}
\end{figure}

\myparagraph{Does Image Difference Level Affect Model Performance?} 
Table~\ref{tab:difference_level} evaluates VLMs across two image difference levels: \textbf{Slight} and \textbf{Significant}.
Models perform substantially better when image differences are significant, as evidenced by notably higher accuracies.
This indicates that VLMs rely heavily on superficial cues rather than capturing the fine-grained visual differences.

\myparagraph{What is a Typical Failure Case?}
Figure~\ref{fig:error_analysis} shows example results, with paired predictions for OSI in blue and IR in red, and a single prediction for SCI in orange.
Incorrect predictions are marked in gray, while cases where all predictions are correct are indicated in green.
As shown, in the OSI and IR tasks, VLMs frequently output biased answers irrespective of the input, predicting both items as either ``A'' or both as ``B''.
For instance, given a pair where $\bm{t}^{a}$ is ``A: a closed cabinet'' and $\bm{t}^{b}$ is ``B: an open cabinet'', many VLMs output ``B'' for both. 
This type of error accounts for 92.67\% of incorrect predictions.
Below, we analyze this behavior for subtle difference discrimination.

\myparagraph{Is the Difficulty Rooted in the Embedding Stage?} One critical question raised by these failure cases is whether subtle visual differences become indistinguishable as early as the embedding stage of model processing.
To examine this possibility, we divided the test examples within the OSI task into correct and incorrect prediction groups using Qwen2.5VL-Inst,
and computed the average cosine similarity between the image embeddings (outputs of the vision transformer encoder) for image pairs $\bm{x}^{a}_{i}$ and $\bm{x}^{b}_{i}$.
This resulted in average similarities of $0.433 \pm 0.052$ (correct) and $0.447 \pm 0.061$ (incorrect), respectively.
Although incorrect cases exhibit higher similarity, these results alone do not indicate a complete loss of visual discriminability at the embedding stage.
Rather, we hypothesize that discriminative visual information is still preserved within the embeddings.

\myparagraph{Is the Subtle Visual Difference Lost in Decoder Layers?} A follow-up question is whether the subtle distinctions become indistinguishable within certain layers of the decoder.
To investigate this, we measure the average cosine similarity of hidden features across each decoder layer in Figure~\ref{fig:similarity} (top).
Similar to the embedding state, incorrect cases exhibit consistently higher similarity at every layer, but no single layer shows a distinct spike.
Moreover, examining the difference in similarity in Figure~\ref{fig:similarity} (bottom) reveals that the similarity of incorrect cases gradually approaches that of correct ones as decoding proceeds through the decoder layers.
Consequently, after decoding, identifiable features likely remain intact regardless of whether the model’s prediction is ultimately correct.

\myparagraph{Most Failures Are Attributable to the Linear Head} Finally, our analysis shows that in 73.76\% of test cases, the model captures subtle visual differences up to the layer immediately preceding the final linear head.
Consider again the common error scenario described above (\textit{i.e.}, predicting the same label for both images).
In the case where both predictions are ``A'', the output token probabilities satisfy $p(\text{A}|\bm{x}^{a}) > p(\text{B}|\bm{x}^{a})$ and $p(\text{A}|\bm{x}^{b}) > p(\text{B}|\bm{x}^{b})$. When we flip one label by assigning the label ``A'' to the image with the higher $p(\text{A}|\cdot)$ and vice versa, $\mathrm{RAcc}$ finally improves significantly to 73.76\%.
Although this strategy diverges from the STATUS Bench protocol, it strongly suggests that the VLM preserves sufficient information to distinguish subtle differences all the way up to the final linear head.
We detailed this analysis in Appendix.

%% file: sections/05_conclusion.tex
\section{Conclusion}
We introduced \textbf{STATUS Bench}, the first benchmark for rigorously evaluating how well VLMs can understand object states.
STATUS Bench is built upon quintuplet data and requires VLMs to solve three tasks (OSI, IR, and SCI) simultaneously.
This evaluation scheme makes it possible to measure not only individual task accuracy but also overall consistency in object state recognition.
Furthermore, we presented \textbf{STATUS Train}, a diverse and large-scale dataset of 13 million semi-automatically generated samples focused on object states and their transitions.
Through extensive experiments with state-of-the-art VLMs, we found that many open-weight VLMs achieve chance-level zero-shot performance in terms of ROA, suggesting that subtle visual information appears to remain discernible in the intermediate embeddings.
We also demonstrated that fine-tuning on STATUS Train can improve accuracy to levels comparable to closed-source models like Gemini 2.0 Flash.
As our primary focus in this study is to propose a new benchmark, we leave improvements in fine-tuning methods and the new design of encoder architectures for handling multiple images as future work.
We hope our STATUS Bench serves as a catalyst for advancing VLMs' capabilities in recognizing object states, making them even better collaborators with humans.

%% file: sections/99_supplemental.tex
\clearpage

\setcounter{page}{1}

\appendix

\begin{center}
    {\Huge Appendix}
\end{center}

\section{Detailed Analysis with Token Probabilities}

To further investigate the role of the final linear head in model failures across multiple VLMs, we conducted a detailed analysis.
Specifically, for both OSI and IR tasks, we examined the output token probabilities produced by VLMs.
By reassigning labels based on these probabilities, \textit{i.e.,} assigning the label with higher output token probability, we observed significant improvements in RAcc metrics as shown in Table~\ref{tab:res_token_prob}.
As discussed in Section 5.3, this finding suggests that VLMs indeed preserve sufficient information to distinguish subtle visual differences all the way up to the final linear head.

\section{Number of Samples in Each Split}

Table \ref{tab:samples_in_split} presents the number of samples in each category and difference level. The \textbf{Positional} state refers to changes caused by movement, whereas the \textbf{Functional} state involves changes resulting from turning on or off. \textbf{Others} includes state related to changes such as cutting and mixing.
In addition, we divide the samples into \textbf{Slight} and \textbf{Significant} levels based on the degree of visual difference between images.

\section{Statistical Analysis on STATUS Train}

Table 7 shows the category-wise distribution of samples for both STATUS Bench and STATUS Train. The STATUS Bench dataset includes all samples, while STATUS Train was calculated by randomly sampling 500 samples. As shown in the table, although the category proportions differ slightly between the two datasets, each category is represented by a sufficient proportion of samples.

\section{Prompts for Inference}

\begin{figure}[t]
    \centering
    \includegraphics[width=\linewidth]{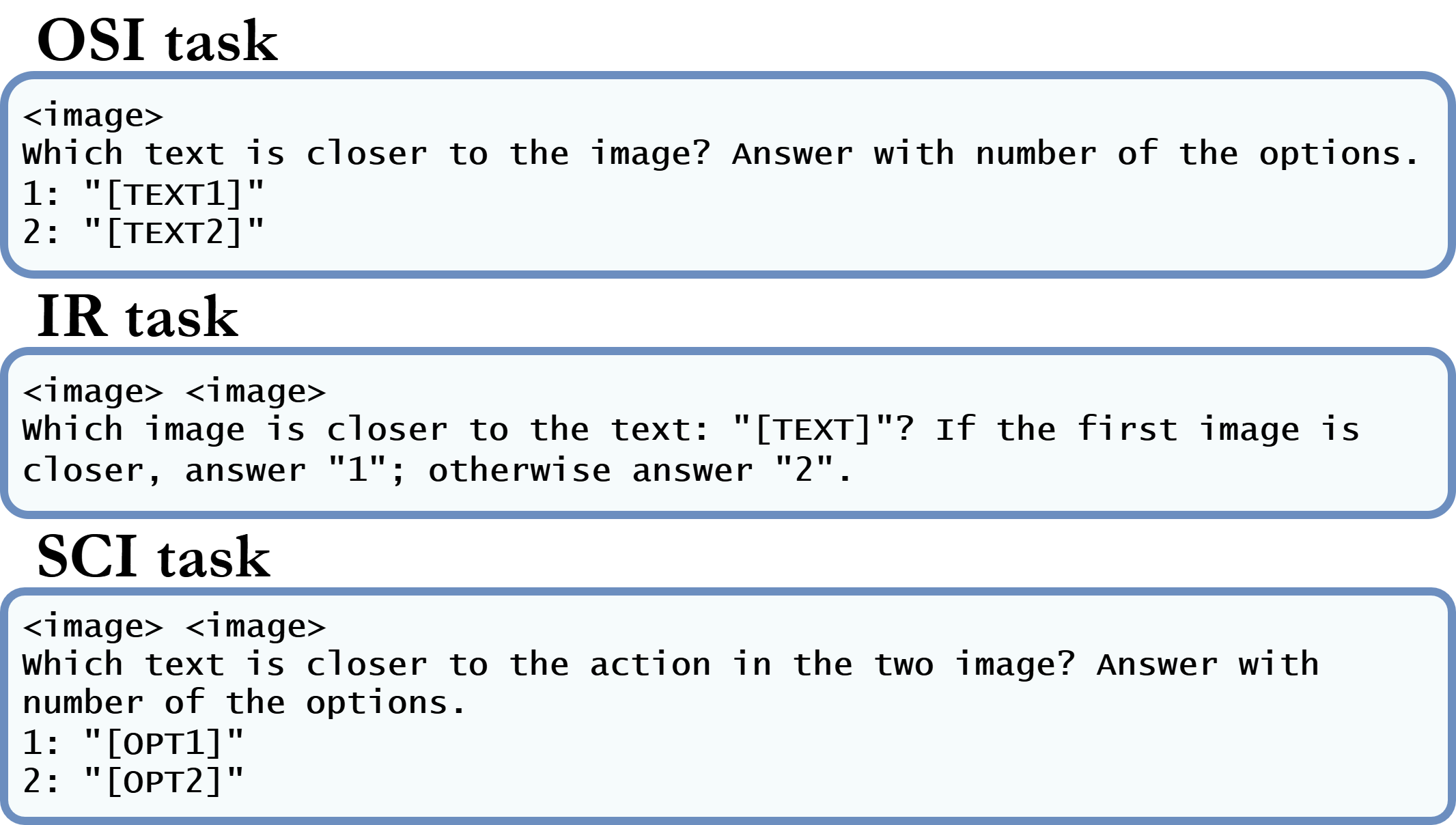}
    \caption{Prompts used in experiments.}
    \label{fig:prompts}
\end{figure}

We show prompts for the experiments in Figure \ref{fig:prompts}. Minor modifications were made to the prompts for each VLM, as different models require different methods for image input, and some did not strictly follow the given instructions.

\section{Computational Resources}

We used two NVIDIA H100 Tensor Core GPUs for LoRA training. The experiment took approximately one week. While full fine-tuning with the same models would require about a month using four NVIDIA H100 Tensor Core GPUs, such a setup was not feasible given our available computational resources.

\newpage

\begin{table}[t]
\centering
\setlength{\tabcolsep}{7.2pt}
\caption{Results obtained by reassigning labels based on output token probabilities assigned by VLMs.}
\label{tab:res_token_prob}
\begin{tabular}{l|cc}
\toprule
\textbf{Model}
& $\bm{\mathrm{RAcc}_{\mathrm{OSI}}}$ &
$\bm{\mathrm{RAcc}_{\mathrm{IR}}}$ 
\\
\midrule
NVILA~\cite{liu2024nvila} & 72.28 & 44.06 \\
Llama-3.2-Vision-Inst~\cite{grattafiori2024llama3} & 53.47 & 50.00  \\
VideoLLaMA3~\cite{cheng2024videollama2} & 52.23 & 53.96 \\
Intern2.5VL~\cite{chen2025intern25vl} & 59.65 & 50.99 \\
LLaVA-OneVision~\cite{li2024llavaonevision} & 77.48 & 74.50 \\
\rowcolor[HTML]{ffede6} LLaVA-OneVision (fine-tuned) & 77.72 & 75.25 \\
Qwen2.5VL-Inst~\cite{bai2025qwen25vl} & 73.27 & 69.55 \\
\rowcolor[HTML]{ffede6} Qwen2.5VL-Inst (fine-tuned) & 73.76 & 71.53 \\
\bottomrule
\end{tabular}
\end{table}

\begin{table}[t]
\centering
\caption{Number of samples in each state category and difference level.}
\label{tab:samples_in_split}
\begin{tabular}{c|c|c||c|c}
\toprule
Positional & Functional & Others & Slight & Significant \\
\midrule
161 & 107 & 136 & 157 & 247 \\
\bottomrule
\end{tabular}
\end{table}

\begin{table}[t]
\setlength{\tabcolsep}{3pt}
\centering
\caption{Category statistics on the STATUS Train and 500 samples in the STATUS Bench.}
\label{tab:cat_in_bench_and_train}
\begin{tabular}{l||c|c||c|c|c}
\toprule
Datasets & Positional & Functional & Others & Slight & Significant  \\
\midrule
Bench & 40.2 & 26.8 & 34.0 & 38.8 & 61.2
\\
\midrule
Train & 43.0 & 30.2 & 26.8 & 30.4 & 69.6 \\
\bottomrule
\end{tabular}
\end{table}

